\documentclass[twoside,twocolumn]{article}

\usepackage{fitee}
\usepackage{kz_style}

\usepackage[hidelinks, breaklinks = true]{hyperref}  % for DVI to PDF latex
\addto{\captionsenglish}{%
  
}
 
\begin{document}

%% article title
\title{Decentralized Multi-Agent Reinforcement Learning \\with Networked Agents: Recent Advances}

%% when all authors provide emails, no $\dagger$
\author[$\ddagger$1]{Kaiqing Zhang}%
\author[2]{Zhuoran Yang}
\author[1]{Tamer Ba\c{s}ar}%
\affil[1]{Coordinated Science Laboratory, University of Illinois at Urbana-Champaign, IL, USA}
%, 1308 West Main, Urbana, IL 61801, USA}
\affil[2]{Department of Operations Research and Financial Engineering, Princeton University, NJ, USA}

\shortauthor{Zhang et al.}	% one author: only Zhai; two authors: Zhai and Hu; three authors: Zhai et al.
 
\authmark{}

%% Only 1 affiliations
%\author{Wen-fei WANG}
%\author{Rong XIONG$^{\dagger\ddagger}$}
%\author{Jian CHU}
%\affil{\it\footnotesize State Key Laboratory of Industrial Control Technology, Institute of Cyber-Systems and Control,
%	\authorcr\it\footnotesize Zhejiang University, Hangzhou 310027, China}

% for authors, use \authorcr to begin with a new line, e.g., \author[2]{\authorcr Xian-liang HU}
% for the affiliation, use \authorcr\Affilfont\it to begin with a new line, e.g.,
%\affil[1]{Editorial Office of Journal of Zhejiang University SCIENCE,
%\authorcr\Affilfont\it Hangzhou 310027, China}

\corremailA{kzhang66@illinois.edu} 
\corremailB{zy6@princeton.edu}
\corremailC{basar1@illinois.edu} 
%\corremailC{abc@zju.edu.cn}
%\corremailD{ABC@zju.edu.cn}
%\emailmark{$\dagger$}	% when all authors provide emails, no \dagger

% abbrev. of month: Jan. Feb. Mar. Apr. May June July Aug. Sept. Oct. Nov. Dec.
\dateinfo{Received Nov. 30, 2019}
%	Revision accepted mmm.\ dd, 2016;    Crosschecked mmm.\ dd, 2017}

\abstract{Multi-agent reinforcement learning (MARL) has long been a significant and everlasting  research topic in both machine learning and control. With the recent development of (single-agent) deep RL, there is a resurgence of interests in developing new MARL algorithms, especially those that are backed by theoretical analysis. 
In this paper, we review some recent advances a sub-area of this topic: decentralized MARL with networked agents. Specifically, multiple agents perform sequential decision-making in a common environment, without the coordination of any \emph{central controller}. Instead, the agents are allowed to exchange information with their neighbors over a communication network. Such a setting finds broad applications in the control and operation of robots, unmanned vehicles, 
mobile sensor networks, and smart grid. This review is built upon several our research endeavors in this direction, together with some progresses made by other  researchers along the line. We hope this review to inspire the devotion of more research efforts  to this exciting yet challenging area. }

% separate by semicolons 

%\doi{10.1631/FITEE.1000000}	% DOI of the paper, should be accurate
%\code{A}
%\clc{TP}

\inpress	% uncomment this command if use "in press"

\publishyear{2018}
\vol{19}
\issue{1}
\pagestart{1}
\pageend{5}

%% when no funding, the following line should be removed, no period at last
\support{Project supported in part  by   the US Army Research Laboratory (ARL) Cooperative Agreement W911NF-17-2-0196, and in part by the Air Force Office of Scientific Research (AFOSR) Grant FA9550-19-1-0353.
} 

%\conf{A preliminary version of this paper has been presented at ??? Conference, date} 
%\esm{Electronic supplementary materials: The online version of this article (http://dx/doi.org/10.1631/jzus.C1000000) contains supplementary materials, which are available to authorized users}
\orcid{Kaiqing Zhang, http://orcid.org/0000-0002-7446-7581}	% corresponding author, or first author
\articleType{Review}
%\articleType can be `Science Letters:', `Review:', `Comment:', etc.
%Leave blank for research article.

\maketitle

\section{Introduction}\label{sec:introduction}

Reinforcement learning (RL) has achieved tremendous successes recently in  many   sequential  decision-making problems, especially associated with  the development of  deep neural networks  for function approximation \citep{mnih2015human}. Preeminent examples  include  playing the game of Go \citep{silver2016mastering, silver2017mastering}, robotics \citep{kober2013reinforcement,lillicrap2016continuous}, and autonomous driving \citep{shalev2016safe}, etc.  
Most of the applications, interestingly,   involve  more than one single agent/player\footnote{Hereafter, we will interchangeably use \emph{agent} and \emph{player}. }, which naturally fall into the realm of   multi-agent RL (MARL). In particular, MARL models the sequential decision-making of multiple autonomous  agents in a common environment, while each agent's objective and the system evolution are both affected by the joint decision made by all agents.  
%Other significant  application scenarios of MARL  include  finance  \citep{lee2002stock,lee2007multiagent}, social science \citep{castelfranchi2001theory,leibo2017multi}, sensor/communication networks \citep{cortes2004coverage,choi2009distributed}, and cyber-physical systems \citep{adler2002cooperative,wang2016towards}.  
%. Specifically, MARL  addresses the sequential decision-making problem of  multiple autonomous  agents that operate in a common environment, each of which aims to optimize its own long-term return by interacting with the environment and other agents Besides the aforementioned popular ones, learning in multi-agent systems finds potential  applications in other subareas, including cyber-physical systems \citep{adler2002cooperative,wang2016towards}, finance  \citep{lee2002stock,lee2007multiagent}, sensor/communication networks \citep{cortes2004coverage,choi2009distributed}, and social science \citep{castelfranchi2001theory,leibo2017multi}. 
MARL algorithms can be generally categorized into three groups, according to the settings they address:  \emph{fully cooperative}, \emph{fully competitive}, and \emph{a mix of the two}  \citep{bu2008comprehensive,zhang2019multiagent}. Specifically, fully cooperative MARL agents aim to optimize a long-term return that is common to all; while fully competitive MARL agents usually have completely misaligned returns that sum up to zero. Agents in the mixed MARL setting, on the other hand, can be both fully cooperative and competitive. In the present  review, for simplicity, we refer to the first ones as \emph{cooperative} MARL, and the second and third ones as \emph{non-cooperative} MARL, respectively.

  There exist several  long-standing  challenges in both cooperative and non-cooperative MARL,  especially in the theoretical analysis for it. 
  First, since  the agents' objectives may be  misaligned with each other, the learning goals in MARL are not just single-dimensional, introducing the challenge of handling \emph{equilibrium} points, and  several   performance criteria other than return-optimization, e.g.,  the communication/coordination efficiency, and the robustness against potential adversaries.  Second, it is well-known that the environment faced by each agent is \emph{non-stationary} in MARL, as it is affected not only by the underlying system evolution, but also by the decisions made by other agents, who are   concurrently improving their policies.   This non-stationarity   invalidates the  framework of most  theoretical analyses in single-agent RL, which are stationary and Markovian.   Third, since   the joint action space  increases exponentially with the number of agents, MARL algorithms may suffer from the  \emph{scalability issues} by nature.  Fourth, the information structure, which dictates the information availability to each agent, becomes more complicated in multi-agent settings, as some of the observations may not be sharable to each other, and sometimes kept in a decentralized fashion.  
  Therefore, the theoretical analysis of MARL algorithms is still relatively lacking  in the literature.

%There has    been a huge volume of work to address the above   challenges. 

%See \cite{bu2008comprehensive} for a comprehensive overview of earlier theories and algorithms on MARL. 
Besides the earlier works on MARL as summarized in  \cite{bu2008comprehensive},  there has been a resurgent interest in this area, especially with the advances of single-agent RL recently \citep{foerster2016learning,zazo2016dynamic,gupta2017cooperative,lowe2017multi,omidshafiei2017deep,zhang2018fully}. Most of these works, with deep neural networks  for function approximation, are not placed  under rigorous theoretical footings, due to the limited understanding of even single-agent deep RL theories. On the other hand, a relatively new paradigm for MARL, \emph{decentralized MARL with networked agents}, has gained increasing research attention \citep{kar2013cal,zhang2018fully,wai2018multi,doan2019finitea}. This is partly due to the fact that  the algorithms under this paradigm require no existence of any central controller, i.e., can be implemented in a decentralized fashion. This can partially address the scalability issues, one of the   aforementioned challenges, and more amenable to a decentralized information structure that is common in practical multi-agent systems \citep{rabbat2004distributed,corke2005networked,dall2013distributed}. The second reason for its popularity is that most algorithms under this paradigm are accompanied with theoretical analysis for convergence/sample complexity, as they are closely related to, and inspired by the recent development of \emph{distributed/consensus optimization} with networked agents, across the areas of control \citep{nedic2009distributed}, operations research \citep{nedic2017achieving}, signal processing \citep{sayed2014adaptation,shi2015extra}, and statistical learning \citep{boyd2011distributed,fan2015multi}.

Specifically, we focus on the MARL setting where the agents, mostly cooperative, are connected by a communication network for the information exchange with each other.  The setting is decentralized in the sense that each agent makes their own decisions, based on only \emph{local} observations and information transmitted from its \emph{neighbors}, without the coordination of any central controller. Such a setting finds broad applications in practice, such as robotics \citep{corke2005networked}, unmanned vehicles \citep{qie2019joint}, mobile sensor networks   \citep{rabbat2004distributed}, intelligent transportation systems \citep{adler2002cooperative,zhang2018dynamic}, and smart grid \citep{dall2013distributed,zhang2018dynamic}, which enjoys several advantages over a centralized setting, in terms of  either cost, scalability, or robustness. For example, it might be costly to even establish a central controller for coordination for some systems \citep{adler2002cooperative,dall2013distributed}, which also easily suffers from malicious attacks and high communication traffic, as the malfunctioning of the central controller will take down the overall system as a whole, and the communication  is concentrated at one place, between the controller and the agents.    As a result, it is imperative to summarize the theories and algorithms  on this topic, for the purpose of both highlighting the boundary of existing research endeavors, and   stimulating future research directions. 

%Recently, this  domain has gained resurgence of interest due to the advances of single-agent RL techniques. Indeed, a huge volume of work on MARL has appeared   lately, focusing on either identifying new learning criteria and/or setups \citep{foerster2016learning,zazo2016dynamic,zhang2018fully,subramanian2019reinforcement}, or developing new algorithms for existing setups, thanks to the development of deep learning \citep{heinrich2016deep,lowe2017multi,foerster2017counterfactual,gupta2017cooperative,omidshafiei2017deep,kawamura2017neural,zhang2019monte}, operations  research \citep{mazumdar2018convergence,jin2019minmax,zhang2019policyb,sidford2019solving}, and multi-agent systems \citep{oliehoek2016concise,arslan2017decentralized,yongacoglu2019learning,zhang2019online}.  
%Nevertheless, not all the efforts are placed under rigorous theoretical footings, partly due to the limited understanding of even single-agent deep RL theories, and partly due to the inherent challenges in  multi-agent settings. As a consequence, it is imperative to review and organize the MARL algorithms with theoretical guarantees, in order to highlight the boundary of existing research endeavors, and stimulate potential future directions on this topic.   
%The resurgence is also driven by several new application scenarios, such as Poker AI and 

%Specifically, XXXX
%This provide desired properties....
 
In this paper, we provide such a review  of  recent advances  on decentralized MARL with networked agents, based on our recent review \cite{zhang2019multiagent} on general MARL algorithms .
%Several other reviews on MARL have appeared  lately \citep{hernandez2017survey,hernandez2018multiagent,nguyen2018deep,oroojlooyjadid2019review,zhang2019multiagent}, which can be viewed as complementary   to ours:   in \cite{hernandez2017survey}, the MARL algorithms that are specifically developed to handle \emph{opponent-induced non-stationarity}, one of the challenges in MARL, have been surveyed;  \cite{hernandez2018multiagent} and \cite{nguyen2018deep} have provided relatively comprehensive  reviews, but with focuses on \emph{deep} MARL, with scarce theories summarized;  in \cite{oroojlooyjadid2019review}, only \emph{cooperative} algorithms have been reviewed. In addition, 
Indeed,  
 \cite{zhang2019multiagent} has provided a comparatively complete overview of general MARL algorithms that are backed by theoretical analysis, serving as the big picture and basis of the present review. Interested readers are referred to \cite{zhang2019multiagent} for a more detailed review.    
The present review summarizes several  our earlier works on this decentralized MARL setting   \citep{zhang2018fully,zhang18cdc,zhang2018finite}, together with some  recent progresses by other researchers along the line. We expect our review to provide  continuing  stimulus for  researchers with similar interests  in working on this exciting yet challenging area.

%\vspace{4pt}
%{\noindent \bf Roadmap.}
%%\vspace{2pt}
%The remainder of this paper is structured as follows. The necessary  background of MARL, especially the decentralized setting with networked agents, is introduced in \S\ref{sec:background}, followed by a summary of algorithms in this setting in  \S\ref{sec:algorithms}. Concluding remarks and several future research directions are provided in  \S\ref{sec:conclusion}. 

\section{Background}\label{sec:background}

In this section,  we provide the necessary background on MARL, especially the decentralized setting with networked agents. 

\subsection{Single-Agent RL}\label{subsec:single_RL}
A general RL agent is  modeled to perform sequential decision-making 
%by interacting with the environment. The environment is usually formulated as 
in a Markov decision process (MDP), as formally defined below. 

\vspace{3pt}
\begin{definition} 
	A \emph{Markov decision process} is defined by a tuple $(\cS,\cA,\cP,R,\gamma)$, where $\cS$ and $\cA$ denote the state and action spaces, respectively;  $\cP:\cS\times\cA\to\Delta(\cS)$ denotes the transition probability from any state $s\in\cS$ to any state $s'\in\cS$ for any given action $a\in\cA$; $R:\cS\times\cA\times\cS\to\RR$ is the reward function that determines the immediate reward received by the agent for a transition from $(s,a)$ to $s'$; $\gamma\in[0,1]$ is the discount factor that trades off the instantaneous and future rewards. 
\end{definition}

%As a standard model, MDP has been widely adopted to characterize the decision-making of an agent with \emph{full observability} of the system state $s$.\footnote{The partially observed MDP (POMDP) model is usually advocated when the agent has no access to the exact system state but only an \emph{observation} of the state. See {\cite{monahan1982state,cassandra1998exact}} for more details on the POMDP model.}  
At each time $t$, the agent chooses to execute an action $a_t$ in face of the system state $s_t$, which causes the system to transition to $s_{t+1}\sim \cP(\cdot\given s_t,a_t)$. Moreover, the agent receives an instantaneous  reward $R(s_t,a_t,s_{t+1})$. 
The goal of the agent is to find a policy $\pi:\cS\to\Delta(\cA)$ so that $a_t\sim \pi(\cdot\given s_t)$ maximizes  the discounted accumulated reward 
\$
\EE\bigg[\sum_{t\geq 0}\gamma^t R(s_t,a_t,s_{t+1})\bigggiven a_t\sim\pi(\cdot\given s_t),s_0\bigg]. 
\$ 

%Accordingly, one can define the \emph{state-action function/Q-function}, and \emph{value function}  under policy $\pi$ as
%\$
%Q_\pi(s,a)&=\EE\bigg[\sum_{t\geq 0}\gamma^t R(s_t,a_t,s_{t+1})\bigggiven a_t\sim\pi(\cdot\given s_t),a_0=a,s_0=s\bigg],\\V_\pi(s)&=\EE\bigg[\sum_{t\geq 0}\gamma^t R(s_t,a_t,s_{t+1})\bigggiven a_t\sim\pi(\cdot\given s_t),s_0=s\bigg]
%\$
%for any $s\in\cS$ and $a\in\cA$, which are the discounted accumulated reward starting from $(s_0,a_0)=(s,a)$ and $s_0=s$, respectively. The ones corresponding to the optimal policy $\pi^*$ are usually referred to as the \emph{optimal Q-function} and the \emph{optimal value function}, respectively.  

Due to the Markovian property, the  optimal policy can be calculated by dynamic-programming/backward induction, such as  value iteration and policy iteration  \citep{bertsekas2005dynamic}, which require  the full knowledge of the model.   
Reinforcement learning, on the other hand,  is devised to find the optimal policy without knowing the model, but by learning from  experiences collected by interacting with either the environment or the simulator.   
In general, RL algorithms can be categorized into two types, \emph{value-based} and \emph{policy-based} methods.

\vspace{4pt}
\noindent{\bf Value-Based Methods:} 
%\vspace{3pt}
Value-based methods aim to find an  estimate of the state-action value/Q- function, which leads to the optimal policy by taking the greedy action with respect to the estimate. Classical value-based RL algorithms include Q-learning \citep{watkins1992q} and SARSA \citep{singh2000convergence}. 
Another important task in RL  that is related to value functions is  to estimate the value function of a  \emph{fixed policy}  (not necessarily the optimal one). This task is referred to as \emph{policy evaluation}, and can be addressed by standard algorithms such as \emph{temporal difference} (TD) learning \citep{tesauro1995temporal,tsitsiklis1997analysis} and gradient TD methods  \citep{sutton2008convergent,bhatnagar2009convergent,sutton2009fast,liu2015finite}. 

%, usually referred to as \emph{policy evaluation}, has been tackled by  algorithms that follow a  similar update as \eqref{equ:Q_learning}, named  \emph{temporal difference} (TD) learning \citep{tesauro1995temporal,tsitsiklis1997analysis,sutton2018reinforcement}. Some other common policy evaluation algorithms with convergence guarantees include gradient TD methods  with linear \citep{sutton2008convergent,sutton2009fast,liu2015finite}, and nonlinear function approximations   \citep{bhatnagar2009convergent}. See \cite{dann2014policy} for a more detailed review on policy evaluation. 

\vspace{4pt}
\noindent{\bf Policy-Based Methods:} 
%\vspace{3pt}
Policy-based methods propose to   directly searches for the optimal one over the policy space, while the space is generally parameterized by function approximators like neural networks, i.e.,  parameterizing  $\pi(\cdot\given s)\approx\pi_\theta(\cdot\given s)$.  
Hence, it is straightforward to improve the policy following the \emph{gradient}  direction of the long-term return, known as the policy gradient (PG) method. \citep{sutton2000policy}   has derived the   closed-form of PG as 
\$
\nabla J(\theta)=\EE_{a\sim\pi_\theta(\cdot\given s),s\sim\eta_{\pi_\theta}(\cdot)}\big[Q_{\pi_\theta}(s,a)\nabla\log\pi_\theta(a\given s)\big],
\$
where $J(\theta)$ and $Q_{\pi_\theta}$ are  the  return and Q-function under policy $\pi_\theta$, respectively,   $\nabla\log\pi_\theta(a\given s)$ is the score function of the policy, and $\eta_{\pi_\theta}$ is the state occupancy measure, either discounted or ergodic, under policy $\pi_\theta$.  
Other standard PG methods include   REINFORCE {\citep{williams1992simple}}, G(PO)MDP {\citep{baxter2001infinite}}, actor-critic  \citep{konda2000actor}, and deterministic PGs \citep{silver2014deterministic}. 

%have been proposed by estimating the gradient in different ways.  A similar idea also applies to deterministic policies in continuous-action settings, whose PG  has been derived   by \cite{silver2014deterministic}. 
%Besides gradient-based ones, several other policy optimization methods have achieved state-of-the-art performance in many applications, including PPO  \citep{schulman2017proximal},  TRPO \citep{schulman2015trust}, and soft actor-critic \citep{haarnoja2018soft}.  

%Compared with value-based methods, policy-based ones enjoy better convergence guarantees \citep{konda2000actor,yang2018finite,zhang2019global,agarwal2019optimality}, especially with neural networks for function approximation \citep{liu2019neural,wang2019neural}, which can readily handle massive or even continuous state-action spaces.  
%Besides the value- and policy-based methods, there also exist RL algorithms based on the linear program formulation of an MDP; see recent efforts in \citep{chen2016stochastic,wang2017primal}. 

\subsection{Multi-Agent RL Framework}\label{subsec:MARL_framework}

Multi-agent RL  also addresses the  sequential decision-making problems, but with more than one agent involved. Specifically,  both the system state evolution  and the reward received by each agent are influenced by the joint actions of all agents. Moreover, each agent has its own long-term reward to optimize, which now becomes a function of the  policies of all other agents. Though various MARL frameworks exist in the literature \citep{bu2008comprehensive,zhang2019multiagent}, we here focus on two examples that are either representative or pertinent to our decentralized MARL setting. 
%Such a general model finds broad applications in practice, see \S\ref{sec:applications} for a detailed review of several significant ones. 

%In general, there exist two seemingly different but closely related theoretical frameworks for  MARL, Markov/stochastic games and extensive-form games, as to be introduced next. Evolution of the systems under different frameworks are illustrated in Figure \ref{fig:model}. 

\vspace{4pt}
{\noindent \bf Markov/Stochastic Games:} As a direct generalization of MDPs to the multi-agent setting, Markov games (MGs),   also known as stochastic games \citep{shapley1953stochastic} has long been treated as a classical framework of MARL \citep{littman1994markov}.  
%%\vspace{2pt}
%
%\subsubsection{Markov/Stochastic Games}\label{subsec:Markov_Games}
%%\issue{2019.08.31.}
%One direct generalization of MDP that captures the intertwinement of multiple agents is Markov games (MGs),   also known as stochastic games \citep{shapley1953stochastic}. 
%Originated from the seminal work \cite{littman1994markov}, the framework of MGs has long been used in the literature  to develop   MARL algorithms, see \S\ref{sec:algorithms} for more details.
%We introduce the  
A formal definition of MGs is introduced as  follows. 

\vspace{3pt}
\begin{definition}\label{def:Markov_Game}
	A \emph{Markov game}  is defined by a tuple $(\cN,\cS,\{\cA^i\}_{i\in\cN},\cP,\{R^i\}_{i\in\cN},\gamma)$, where $\cN=\{1,\cdots,N\}$ denotes the set of  $N> 1$ agents, 
	$\cS$ denotes the state space observed by all agents, $\cA^i$ denotes the action space of agent $i$. Let $\cA:=\cA^1\times\cdots\times\cA^N$, then 
	 $\cP:\cS\times\cA\to\Delta(\cS)$ denotes the transition probability from any state $s\in\cS$ to any state $s'\in\cS$ for any joint action $a\in\cA$; $R^i:\cS\times\cA\times\cS\to\RR$ is the reward function that determines the immediate reward received by agent $i$ for a transition from $(s,a)$ to $s'$; $\gamma\in[0,1]$ is the discount factor. 
\end{definition}

At time $t$, each agent $i\in\cN$ chooses an action $a^i_t$, according to the system state $s_t$. The joint chosen action $a_t=(a^1_t,\cdots,a^N_t)$ then makes the  system  transition to state $s_{t+1}$, and  assigns to each agent $i$ a reward  $r^i_t=R^i(s_t,a_t,s_{t+1})$. Agent $i$'s goal  is to finding the policy $\pi^i:\cS\to\Delta(\cA^i)$ such that its own long-term return is optimized. 
%, by finding the policy $\pi^i:\cS\to\Delta(\cA^i)$ such that $a^i_t\sim \pi^i(\cdot\given s_t)$. 
%a mapping from the state space to the probability over its own action space $\cA^i$. 
Accordingly, the agent $i$'s value-function $V^i:\cS\to\RR$  becomes a function of the joint policy $\pi:\cS\to\Delta(\cA)$ with   $\pi(a\given s):=\prod_{i\in\cN}\pi^i(a^i\given s)$, which is defined as  
\#\label{eq:V_pi}
V^i_{\pi^i,\pi^{-i}}(s):=\EE_{a^i_t\sim \pi^i(\cdot\given s_t)}\bigg[\sum_{t\geq 0}\gamma^t r^i_t\bigggiven s_0=s\bigg],
\#
where $-i$ represents the indices of all agents in $\cN$ except agent $i$. 
Owing to this coupling of polices, 
%With a random initialization $s_0\sim\cD$ for some distribution  $\cD\in\Delta(\cS)$, the 
%Hence, 
the solution concept of MGs is not simply an \emph{optimum}, but an \emph{equilibrium} among all agents.  The most common one, named \emph{Nash equilibrium (NE)} in MGs, is defined as below \citep{basar1999dynamic}. 

%{\color{red} Maybe use $\eta$ for visitation measure.  consistent with EFG}

\vspace{3pt}
\begin{definition}\label{def:NE}
	A \emph{Nash equilibrium} of the MG $(\cN,\cS,\{\cA^i\}_{i\in\cN},\cP,\{R^i\}_{i\in\cN},\gamma)$ is a joint policy $\pi^*=(\pi^{1,*},\cdots,\pi^{N,*})$, such that for any $s\in\cS$  and $i\in\cN$
	\$
	V^i_{\pi^{i,*},\pi^{-i,*}}(s)\geq V^i_{\pi^i,\pi^{-i,*}}(s),\quad\text{~~for any~~}\pi^i.
	\$
\end{definition}

Nash equilibrium describes an   point $\pi^*$, from which no agent has any   incentive to deviate. 
%In other words, for any agent $i\in\cN$, the policy $\pi^{i,*} $ is the \emph{best-response} of $\pi^{-i,*}$. As a standard learning goal for MARL, NE always exists for discounted MGs \citep{filar2012competitive}, but 
%may not be unique in general.  
Most of the MARL algorithms are contrived to converge to such an equilibrium point, making MGs the most standard framework in MARL. 

%XXX explanaiton of NE XXX 

%Discuss different settings, problem formulation, and solution concept:

Indeed, this framework of MGs is general enough to cover both cooperative and non-cooperative MARL settings. For the formal one, all agents share a    common reward function, i.e., $R^1=R^2=\cdots=R^N=R$. Such a model is also known as \emph{multi-agent MDPs} (MMDPs) \citep{boutilier1996planning,lauer2000algorithm} and \emph{Markov teams} \citep{wang2003reinforcement,mahajan2008sequential}. In this setting,  
%We note that this model is also  referred to as \emph{multi-agent MDPs} (MMDPs)  in the  AI community \citep{boutilier1996planning,lauer2000algorithm}, and \emph{Markov teams/team Markov games}   in the control/game theory community \citep{yoshikawa1978decomposition,ho1980team,wang2003reinforcement,mahajan2008sequential}. Moreover, from the game-theoretic perspective, this cooperative setting can also be viewed as a special case of Markov \emph{potential} games \citep{gonzalez2013discrete,zazo2016dynamic,valcarcel2018learning}, with the potential function being the common accumulated reward. 
%With this model in mind, 
the value functions are identical to all agents, enabling the use of 
single-agent RL algorithms, provided that  all agents are coordinated as one decision maker.  The latter setting with non-cooperative agents correspond to the MGs with either \emph{zero-sum} or \emph{general-sum} reward functions. Such {misaligned}  objectives of self-interested agents necessitate  the use of Nash equilibrium as the solution concept. 
% The global optimum for cooperation now constitutes a Nash equilibrium of the game.  

\vspace{4pt}
\noindent{\bf Networked MMDPs:} 
%\vspace{3pt}
As a generalization of the  above common-reward cooperative model, the following one of \emph{networked MMDPs} plays an essential role in decentralized MARL with networked agents.

\vspace{3pt}
\begin{definition}\label{def:networked_MMDP}
	A \emph{networked MMDP}  is defined by a tuple $(\cN,\cS,\{\cA^i\}_{i\in\cN},\cP,\{R^i\}_{i\in\cN},\gamma,\{\cG_t\}_{t\geq 0})$, where the first six elements are identical to those in {Definition \ref{def:Markov_Game}}  for MGs, and $\cG_t=(\cN,\cE_t)$  denotes the time-varying communication network that connects all agents, with $\cE_t$ being the set of communication links at time $t$, i.e., an edge $(i,j)$ for agents $i,j\in\cN$ belongs to $\cE_t$ if agent $i$ and $j$ can communicate with each other at  time $t$. 
\end{definition}

The system evolution of  networked MMDPs is identical to MGs, but with one difference  in terms of the objective: all agents aim to cooperatively optimize the  long-term return  corresponding to the team-average reward $\bar{R}(s,a,s'):=N^{-1}\cdot \sum_{i\in\cN}R^i(s,a,s')$ for any $(s,a,s')\in\cS\times\cA\times\cS$. 
Moreover, each agent makes decisions using only the local information, including the information transmitted from its neighbors over the network.  
The networked MMDP model allows 
% for cooperative  MARL stems from \citep{kar2013cal}
%another slightly more general and surging model for cooperative MARL considers  \emph{team-average}  reward \citep{kar2013cal,zhang2018fully,doan2019finitea}. Specifically, 
agents   to cooperate, but with different  reward functions/preferences.  
% is still common: to optimize  the long-term return  corresponding to the team-average reward $\bar{R}(s,a,s'):=N^{-1}\cdot \sum_{i\in\cN}R^i(s,a,s')$ for any $(s,a,s')\in\cS\times\cA\times\cS$. 
This model is able to not only  capture   more  heterogeneity and privacy among agents (compared to conventional  MMDPs), but also   facilitate  the development of  \emph{decentralized} MARL algorithms with only neighbor-to-neighbor communications \citep{kar2013cal,zhang2018fully,wai2018multi}. 
In addition,  
such heterogeneity also necessitates  the incorporation of more \emph{efficient}  {communication} protocols  into MARL, an  important while relatively open problem in MARL that naturally arises in networked MMDPs \citep{chen2018communication,ren2019communication,lin2019communication}. 

\section{Algorithms}\label{sec:algorithms}

This section  provides a review of MARL algorithms under the frameworks introduced in \S\ref{subsec:MARL_framework}.  Specifically, we categorize the algorithms, which are amenable to the decentralized setting with networked agents, according to the tasks they address, such as learning the optimal/equilibrium policies, and policy evaluation. Besides, we will also mention several algorithms aiming  to achieve other learning goals in this setting. 

%selective  review of MARL algorithms, and categorizes them according to the tasks to address.  Exclusively, we here  review  the works that are focused on the theoretical studies only, which are mostly built upon the two representative MARL frameworks, fully observed Markov games and extensive-form games, introduced in \S\ref{subsec:MARL_framework}. A brief summary  on MARL for partially observed Markov games in  \emph{cooperative}  settings, namely, the Dec-POMDP problems, is also provided below in \S\ref{subsec:cooperative_MARL_alg},  due to their relatively  maturer  theory than that of MARL for general partially observed Markov games.
 
\subsection{Learning Policies}\label{subsec:learning_policy}

We first review the algorithms for the task of \emph{control} in RL, namely, learning the optimal/equilibrium polices for the agents. Algorithms for both cooperative and non-cooperative settings exist in the literature.

\vspace{4pt}
\noindent{\bf Cooperative Setting:} 

Consider a team of agents cooperating under the framework of networked MMDPs introduced in Definition \ref{def:networked_MMDP}.  
Including the framework of MMDPs/Markov teams as a special case, this one  generally requires more coordination, since the  global value function  cannot be estimated locally without knowing the other agents' reward functions. This challenge becomes more severe when  no  central controller, but only neighbor-to-neighbor communication over a network, is available for coordination. 

%, most MARL algorithms reviewed in   \S\ref{subsubsec:homo_agents}  directly apply, since the controller can collect and average the rewards, and distributes the information to all agents. Nonetheless,  such a controller may not exist in 
%most aforementioned applications,  due to  either cost, scalability, or robustness concerns \cite{rabbat2004distributed,dall2013distributed,zhang2018distributedauto}. 
%Instead,   the agents may be able to share/exchange information with their neighbors over  a possibly time-varying and sparse communication network, as  illustrated in    Fig. \ref{fig:info_struc} (b). 

Such an information structure has appeared frequently in the proliferate studies on \emph{decentralized/distributed}\footnote{Note that hereafter  we use \emph{decentralized} and \emph{distributed} interchangeably to  describe this structure, to respect some conventions from distributed optimization literature. } algorithms, such as average consensus \citep{xiao2007distributed} and distributed/consensus optimization \citep{nedic2009distributed,shi2015extra}. Nevertheless, relatively fewer efforts  have devoted to address this structure in MARL.   In fact, most existing results in distributed/consensus optimization can be viewed as  solving \emph{static/one-stage} decision-making problems \citep{nedic2009distributed,agarwal2011distributed,jakovetic2011cooperative,tu2012diffusion}, which is easier to analyze compared to RL, a sequential decision-making setting where the decisions made at current time will have a long-term effect.

%
%\vspace{6pt}
%\noindent{\bf Learning Optimal Policy}
%\vspace{3pt}

%The most significant goal is to learn the optimal joint policy, while each agent only accesses to the local and neighboring information over the network.  
The idea of decentralized  MARL over networked agents dates back to \cite{varshavskaya2009efficient}, for the control of distributed robotic systems.  The algorithm therein uses the idea of average consensus, and is policy-based. However, no theoretical analysis is provided in the work. 
To the best of our knowledge,  under this setting, 
the first MARL algorithm with provable  convergence guarantees is \cite{kar2013cal}, which combines  the idea of \emph{consensus + innovation} \citep{kar2013consensus} to the standard Q-learning algorithm, leading to the \emph{$\mathcal{QD}$-learning} algorithm that is updated as follows:
\small
\$
&Q^i_{t+1}(s,a)\leftarrow Q^i_{t}(s,a)+\alpha_{t,s,a}\Big[R^i(s,a)+\gamma\min_{a'\in\cA}Q^i_t(s',a')\\
&\qquad~~\qquad-Q^i_t(s,a)\Big]-\beta_{t,s,a}\sum_{j\in\cN^i_t}\big[Q^i_t(s,a)-Q^j_t(s,a)\big],
\$
\normalsize
where $\alpha_{t,s,a},~\beta_{t,s,a}>0$ denote the  stepsizes, $\cN^i_t$ denotes agent $i$'s  set of neighboring agents, at time $t$.  Compared to the Q-learning update \citep{watkins1992q}, \emph{$\mathcal{QD}$}-learning adds an \emph{innovation} term, which is the difference between the agent's Q-value estimate and its neighbors'. 
Under some standard conditions on the  stepsizes, $\mathcal{QD}$-learning is proven to converge to the optimum Q-function, for the tabular setting with finite state-action spaces.  

As the joint action space increases exponentially with the number of agents,   function approximation becomes especially  pivotal to the scalability of MARL algorithms. To establish convergence analysis in the function approximation regime, we have resorted to  policy-based  algorithms, specifically, actor-critic algorithms, for this setting \citep{zhang2018fully}.  
%proposes actor-critic algorithms for this setting. 
Specifically, each agent $i$'s  policy is parameterized as $\pi^i_{\theta^i}:\cS\to\cA^i$ by some $\theta^i\in\RR^{m^i}$, and the joint policy is thus defined as $\pi_\theta(a\given s):=\prod_{i\in\cN}\pi^i_{\theta^i}(a^i\given s)$. 
Let $Q_{{\theta}}$ be the global value function corresponding to the team-average reward $\bar R$ under the joint policy $\pi_\theta$. Then, we first establish the   policy gradient of the return w.r.t. each agent $i$'s parameter $\theta^i$ as  
%\small
\#\label{eq:policy_gradient_thm}
		\nabla _{\theta^{i} } J({\theta}) &= \EE \left [ \nabla _{\theta^i} \log \pi _{\theta^i} ^i(s,a^i)\cdot  Q_{{\theta}}(s,a) \right ]. 
%		=\EE  \left [ \nabla _{\theta^i} \log \pi _{\theta^i}^i (s,a^i)\cdot  A_{{\theta}}(s,a) \right ],
%		\nonumber\\
%		&=\EE _{s \sim d_{{\theta}}, a \sim \pi_{\theta} } \left [ \nabla _{\theta^i} \log \pi _{\theta^i}^i (s,a^i)\cdot  A^i_{{\theta}}(s,a) \right ].
\#
\normalsize 
%where  $Q_{{\theta}}$ is the global value function corresponding to $\bar R$ under the joint policy $\pi_\theta$ that is defined as $\pi_\theta(a\given s):=\prod_{i\in\cN}\pi^i_{\theta^i}(a^i\given s)$. 
Analogous  to the single-agent PG given in \S\ref{subsec:single_RL}, the PG in \eqref{eq:policy_gradient_thm}  involves the expectation of the product between the \emph{global}  Q-function $Q_{{\theta}}$, and the \emph{local} score function $\nabla _{\theta^i} \log \pi _{\theta^i} ^i(s,a^i)$.  The former quantity,  however, cannot be estimated locally at each agent. Therefore, by parameterizing each local copy of $Q_{{\theta}}(\cdot,\cdot)$ as $Q_{{\theta}}(\cdot,\cdot;\omega^i)$, a consensus-based TD learning update is proposed for the  \emph{critic}  step, i.e., for estimating $Q_{{\theta}}(\cdot,\cdot)$ given $\pi_\theta$: 
\#
\tilde{\omega}^i_{t}&=\omega^i_{t}+\beta_{\omega,t}\cdot \delta^i_{t}\cdot \nabla_{\omega} Q_t(\omega^i_t),\label{equ:MARL_critic_0}\\
\qquad\qquad{\omega}^i_{t+1}&= \sum_{j\in\cN}c_t(i,j)\cdot \tilde{\omega}^j_{t}
\label{equ:MARL_critic_1},
\#
where $\beta_{\omega,t}>0$ denotes  the stepsize,  and 
$\delta^i_{t}$ is the local TD-error calculated using $Q_{{\theta}}(\cdot,\cdot;\omega^i)$. \eqref{equ:MARL_critic_0} is the  standard TD learning update at agent $i$, while \eqref{equ:MARL_critic_1} is a  weighted  combination step of the neighbors' estimates $\tilde{\omega}^j_{t}$. The weights $c_t(i,j)$ are  determined by the topology of the communication network, namely,  it only has  non-zero values if the two agents $i$ and $j$ are connected at time $t$, i.e., $(i,j)\in\cE_t$. The weights  also need to satisfy the \emph{doubly stochastic} property in expectation, so that $\omega^i_{t}$ reaches a \emph{consensual} value for all $i\in\cN$ as $t\to\infty$.  
Then,  in the actor step, 
each agent $i$ updates its policy following stochastic policy gradient   \eqref{eq:policy_gradient_thm}, using its own Q-function estimate $Q_{{\theta}}(\cdot,\cdot;\omega^i_t)$.  
In addition, motivated by the fact that the temporal difference can also be used in policy gradient to replace the Q-function \citep{bhatnagar2009natural}, we also propose a variant algorithm that relies on not the Q-function, but the state-value function approximation  \citep{zhang2018fully}, in order to reduce the variance in the PG update.  

When linear functions are  used for value function approximation, we can establish the  almost sure convergence of the decentralized actor-critic updates \citep{zhang2018fully}. The proof techniques therein are based on the two-timescale stochastic approximation approach in \cite{borkar2008stochastic}. 
% for these decentralized actor-critic algorithms, when     linear functions are  used for value function approximation. 
Later in  \cite{zhang18cdc}, we extend the 
similar ideas  to the setting specifically with \emph{continuous} spaces, where deterministic policy gradient (DPG) method  is usually used. 
For DPG methods, 
off-policy exploration using a stochastic behavior policy is required in general, as the deterministic on-policy may not be explorative enough. Nonetheless, as the  policies of other agents are unknown in the multi-agent setting, the standard  off-policy approach \citep[\S 4.2]{silver2014deterministic} is not applicable. 
As a result, we develop an actor-critic algorithm \citep{zhang18cdc}, which is still on-policy, using the recent development of  the expected policy gradient (EPG) method \citep{ciosek2018expected}. EPG unifies stochastic PG (SPG) and DPG, but reduces the variance of general SPGs.  Specifically, the critic step remains identical to \eqref{equ:MARL_critic_0}-\eqref{equ:MARL_critic_1}, while the actor step is replaced by the  multi-agent version of EPG we newly derived.  
When linear function approximation is used, we can also establish the almost sure 
convergence of the algorithm. 
%is then also guaranteed when linear function approximation is used \cite{zhang18cdc}.
 In the same vein, the extension of \cite{zhang2018fully} to an off-policy  setting has been investigated in 
 \cite{suttle2019multi}, which is built upon the  emphatic temporal differences (ETD) method for the critic \citep{sutton2016emphatic}. 
 Convergence can also be established using stochastic approximation approach, by 
 incorporating the analysis of ETD($\lambda$) \citep{yu2015convergence} into \cite{zhang2018fully}. 
% , almost sure convergence guarantee has also been established. 
In addition, another   off-policy algorithm for the same setting is proposed  in a concurrent work in \cite{zhang2019distributed}. Deviated from the line of works above, 
agents do not share/exchange their estimates of value function. In contrast, the agents' goal is to reach consensus over the global optimal policy estimation. 
This yields a  local critic and a consensus actor update, which also enjoys provably  asymptotic convergence.

We note that aforementioned convergence guarantees are \emph{asymptotic}, namely, the algorithms are guaranteed to converge only as the iteration  numbers go to infinity.
More importantly, 
these convergence results are restricted to the case with linear function approximations. These two drawbacks make it imperative, while challenging, to  quantify the performance when  finite iterations and/or samples are used, and  when nonlinear functions such as deep neural networks are used in practice. 
Serving as an initial step towards the \emph{finite-sample analyses} in this setting with  more \emph{general} function approximation, we study   in \cite{zhang2018finite} the \emph{batch} RL algorithms \citep{lange2012batch} in the multi-agent setting. In particular, we propose decentralized variants of the fitted-Q iteration (FQI) algorithm  \citep{riedmiller2005neural,antos2008fitted}. We focus on FQI as it  motivates the celebrated deep Q-learning algorithm \citep{mnih2015human} that has achieved  great empirical success.  
%We study FQI variants for both the cooperative setting with networked agents, and the competitive setting with two teams of such networked agents (see \S\ref{sec:value_comp} for more details). In the former setting,   a 
All agents collaborate to update the global Q-function estimate iteratively, by fitting nonlinear least squares with the target values as the responses. Let $\cF$ denote  the  function class for Q-function approximation, $\{(s_j,\{a^i_j\}_{i\in\cN},s_{j}')\}_{j\in[n]}$ be the batch transitions dataset of size $n$  available to all agents, and $\{r^i_j\}_{j\in[n]}$ be the local reward samples private to each agent. Then,  the local target value at each agent $i$ is calculated as $y^i_j=r^i_j+\gamma\cdot \max_{a\in\cA}Q^i_t(s_{j}',a)$, where $Q^i_t$ is agent $i$'s Q-function estimate at iteration  $t$. As a consequence,  all agents aim to collaboratively find a common Q-function estimate by  solving
\#\label{equ:FQI_coop}
\min_{ f\in \cF}~~ \frac{1}{N}\sum_{i\in\cN}\frac{1}{2n} \sum_{j=1}^n \bigl [ y^i_j - f(s_j, a_j^1, \cdots,a_j^N) \bigr ]^2.  
\#
As $r^i_j$, and thus $y^i_j$, is only available to agent $i$, the problem in \eqref{equ:FQI_coop} fits in the standard formulation of \emph{distributed/consensus optimization} \citep{nedic2009distributed,agarwal2011distributed,jakovetic2011cooperative,tu2012diffusion,hong2017stochastic,nedic2017achieving}. If  $\cF$ makes  $\sum_{j=1}^n  [ y^i_j - f(s_j, a_j^1, \cdots,a_j^N)  ]^2$ convex for each $i$, then the global optimum can be achieved by the algorithms in these  references. 
For the special case when $\cF$ is a linear function class, this is indeed the case. 
% if $\cF$ is a linear function class. 

 Unfortunately, with only a finite iteration of  distributed optimization algorithms performed at each agent, the agents may not reach exact consensus. This results in  an error in each agent's Q-function estimate, compared with  the actual optimum of \eqref{equ:FQI_coop}. When nonlinear function approximation is used, this error is even more obvious, as the actual global optimum can hardly be obtained in general. By accounting for this  error due to decentralized computation, we derive the \emph{error propagation}  results following those for  the   single-agent batch RL \citep{munos2007performance,munos2008finite,antos2008fitted,antos2008learning,farahmand2010error}, in order to  establish the finite-sample performance of the proposed algorithms.
Specifically, we establish  the dependence of the accuracy of the algorithms output, on   the function class $\cF$, the number of samples within each iteration $n$, and the number of iterations for $t$. 

\vspace{4pt}
\noindent{\bf Non-Cooperative Setting:}

The networked MMDP model can also be considered in a non-cooperative setting, which though has not been extensively studied in the literature. In \cite{zhang2018finite}, we also consider one type of non-cooperative setting, where two teams of networked agents, Teams $1$ and $2$, form a zero-sum Markov game as introduced in Definition \ref{def:Markov_Game}. Such a setting can be viewed as a mixed one with both cooperative (within each team), and competitive (against the opponent team) agents. We then establish finite-sample analysis for a   decentralized  variant of FQI for this setting. 

In particular, by instantiating the definition of Nash equilibrium  in a two-player  zero-sum case, for a given Q-value  $Q(s,\cdot,\cdot):\cA\times\cB\to \RR$,  one can define a \texttt{Value} operator  at any state $s\in\cS$ as 
\small
\$
\texttt{Value} \bigl[ Q(s,\cdot , \cdot ) \bigr ]  = \max_{u \in \Delta(\cA) }\min _{v\in \Delta(\cB) } \EE_{a\sim u,b\sim v}\big[Q(s,a, b)\big], 
 \$
 \normalsize
where $\cA$ and $\cB$ are the joint action spaces, $u$ and $v$ are the one-stage strategy,  of agents in Teams $1$ and $2$, respectively.  
Additionally, if some function $Q$ satisfies the following fixed-point equation for any $s\in\cS$
\small
\#\label{equ:def_Q_game}
Q(s,a,b)=\bar R(s,a,b)+\gamma\cdot \texttt{Value} \bigl[ Q(s,\cdot , \cdot ) \bigr ],
\#
\normalsize
where $\bar R$ is the team-average reward of Team $1$ (thus $-\bar R$ is that of Team $2$), then such a $\texttt{Value} \bigl[ Q(s,\cdot , \cdot ) \bigr ]$ defines the  \emph{value of the game} at any state $s\in\cS$. 
In comparison to the single-agent case, the $\max\min$ operator, instead of the $\max$ one is used to define the optimal/equilibrium value function. 

Therefore, in order to solve the MARL problem in this setting, it suffices to find a good estimate of the Q-function satisfying \eqref{equ:def_Q_game}.  Hence, similarly as the single-team cooperative setting, 
all agents within one team now collaboratively solve for a common Q-function estimate by solving \eqref{equ:FQI_coop}, but replace  the local target value at each agent $i$ by $y^i_j=r^i_j+\gamma\cdot \texttt{Value} \bigl[Q^i_t(s_{j}',a,b)\bigr]$, and the fitting function $f(s_j, a_j)$ by $f(s_j, a_j,b_j)$, a function over the joint action spaces of both teams. Then, such an   optimization problem is solved in a distributed fashion as \eqref{equ:FQI_coop}. Similar error-propagation analysis can be performed in this setting, leading to the finite-sample error bounds of the decentralized FQI algorithm. To the best of our knowledge, this appears to be the first finite-sample analysis for decentralized batch MARL in non-cooperative settings.

%The independent MDP case \cite{macua2017diff}, \cite{pennesi2010distributed} (where all state and action spaces are local, and the  each agent's action does not change the other agents'  transition models.)

\subsection{Policy Evaluation}\label{subsec:policy_eva}

Besides control, a great number of algorithms have been developed to address the policy evaluation task in this decentralized MARL setting. In particular, policy evaluation corresponds to  the critic step of the aforementioned actor-critic algorithms only. With a fixed policy,  this task enjoys a neater formulation,  because  the sampling  distribution now becomes stationary. Moreover, as linear function approximation is commonly used for this task, the   objective is mostly convex. This makes the  finite-time/sample analyses easier, in comparison to many control algorithms with only  {asymptotic convergence guarantees}.   

Specifically, under joint policy $\pi$, suppose each agent parameterizes the value function by $\{V_\omega(s):=\phi^\top(s)\omega:\omega\in\RR^d\}$, where $\phi(s)\in\RR^d$ is the  feature vector at $s\in\cS$, and $\omega\in\RR^d$ is the parameter vector. For notational convenience, let $\Phi:=(\cdots;\phi^\top(s);\cdots)\in\RR^{|\cS|\times d}$,  $D=\diag[\{\eta_\pi(s)\}_{s\in\cS}]\in\RR^{|\cS|\times|\cS|}$ be a diagonal matrix constructed using the   state-occupancy measure $\eta_\pi$, $\bar R^\pi(s)=N^{-1}\cdot \sum_{i\in\cN}R^{i,\pi}(s)$, where $R^{i,\pi}(s)=\EE_{a\sim\pi(\cdot\given s),s'\sim P(\cdot\given s,a)}[R^i(s,a,s')]$, and $P^\pi\in\RR^{|\cS|\times|\cS|}$ with the $(s,s')$ element being $[P^\pi]_{s,s'}=\sum_{a\in\cA}\pi(a\given s)P(s'\given s,a)$. 
The objective of all agents is to jointly minimize the mean square projected Bellman error (MSPBE) associated with the team-average reward, i.e., 
\#\label{equ:MSPBE_joint}
\min_{\omega}\quad\texttt{MSPBE}(\omega):&=\big\|\Pi_{\Phi}\big(V_\omega-\gamma P^\pi V_\omega -\bar R^\pi\big)\big\|^2_D\notag\\
&=\big\|A\omega-b\big\|^2_{C^{-1}},
\#
where   $\Pi_{\Phi}:=\Phi(\Phi^\top D\Phi)^{-1}\Phi^\top D$ is the projection operator onto subspace $\{\Phi\omega:\omega\in\RR^d\}$, $A:=\EE\{\phi(s)[\phi(s)-\gamma \phi(s')]^\top\}$, $C:=\EE[\phi(s)\phi^\top(s)]$, and $b:=\EE[\bar R^\pi(s)\phi(s)]$. 
Using Fenchel duality, and replacing the expectation with samples,   the finite-sum  version of \eqref{equ:MSPBE_joint} can be re-formulated as a distributed saddle-point problem 
\small
\$
\min_{\omega}\max_{\lambda^i}\frac{1}{Nn}\sum_{i\in\cN}\sum_{j=1}^n 2(\lambda^i)^\top A_j\omega -2(b^i_j)^\top \lambda^i-(\lambda^i)^\top C_j\lambda^i, 
%\texttt{MSPBE}(\omega):=\big\|\Pi_{\Phi}\big(V_\omega-\gamma P^\pi V_\omega -\bar R^\pi\big)\big\|^2_D
\$
where $n$ is the data size, $A_j,C_j$ and $b^i_j$ are empirical estimates of $A,C$ and $b^i:=\EE[R^{i,\pi}(s)\phi(s)]$ using sample $j$, respectively.   
The objective above is convex in $\omega$ and concave in $\{\lambda^i\}_{i\in\cN}$. The use of   MSPBE as an objective is standard in multi-agent policy evaluation \citep{macua2015distributed,lee2018primal,wai2018multi,doan2019finitea}, and the idea of  saddle-point reformulation  has been adopted in \cite{macua2015distributed,lee2018primal,wai2018multi,cassano2018multi}. 
%Note that in \cite{cassano2018multi}, a variant of MSPBE, named H-truncated $\lambda$-weighted MSPBE, is advocated, in order to control the bias of the solution deviated from the actual mean square Bellman error minimizer.  

With the formulation \eqref{equ:MSPBE_joint},  \cite{lee2018primal} develops a distributed variant of the gradient TD-based method \citep{sutton2009fast}, and establishes the  asymptotic   convergence  using the ordinary differential equation (ODE) method.  \cite{wai2018multi} proposes  a double averaging scheme that combines the dynamic consensus  \citep{qu2017harnessing} and the SAG algorithm \citep{schmidt2017minimizing}, in order to solve the saddle-point problem  with a linear rate. In  \cite{cassano2018multi},  the idea of   variance-reduction, specifically, AVRG in \citep{ying2018convergence}, has been incorporated into  gradient TD-based policy evaluation. Achieving the same linear rate as \cite{wai2018multi}, three advantages are claimed in \cite{cassano2018multi}: i)  data-independent memory requirement; ii) use of eligibility traces \citep{singh1996reinforcement}; iii) no need for  synchronization  in sampling. More recently, standard TD learning \citep{tesauro1995temporal}, instead of gradient-TD, has been generalized to this MARL setting, with special focuses on  finite-sample analyses, see  \cite{doan2019finitea,doan2019finiteb}.   
By the proof techniques in \cite{pmlr-v75-bhandari18a}, \cite{doan2019finitea} studies the distributed TD($0$) algorithm.   
%is first studied in \cite{doan2019finitea}, using the proof techniques originated in \cite{pmlr-v75-bhandari18a}, which requires 
A projection operation is required on the iterates, and the data samples are assumed  to be independent and identically distributed (i.i.d.). 
Then, following  the recent advance in \cite{pmlr-v99-srikant19a}, \cite{doan2019finiteb} provides 
finite-time performance of the  more general  distributed TD($\lambda$) algorithm, without the need of any  projection or i.i.d. noise assumption.

%Each agent maintains a local copy of $\theta$, 
% \issue{2019.09.22}
%  
% serves as the common cost  function to minimize
%
%
%\issue{XXXX} easier and thus hope to get better results. 

%introduce the algorithms first, including both for \emph{control} and for \emph{policy evaluation} \cite{wai2018multi,cassano2018multi,doan2019finitea,doan2019finiteb},.

%Also, note that several distributed results (over network) assume ``independent MDP'', for policy evaluation \cite{macua2015distributed,stankovic2016distributed,stankovic2016multi}. 

\subsection{Other Learning Goals}\label{subsec:other_learning}
 
Several other learning goals have also been investigated  in this setting.  \cite{zhang2016data} considers  the \emph{optimal consensus}  problem, where each agent   tracks its neighbors' as well as a leader's states, so that the  consensus error is minimized by the joint policy. Then, 
a policy iteration algorithm is devised, and made practical by introducing an actor-critic algorithm with neural networks for function approximation. 
\cite{zhang2018model} also uses 
a similar consensus error objective, with  the name of cooperative multi-agent graphical games. 
Off-policy RL algorithms are developed, using a centralized-critic-decentralized-actor scheme.

As an essential ingredient in the algorithm design for the decentralized MARL settings, communication efficiency in MARL  has drawn increasing attention recently \citep{chen2018communication,ren2019communication,lin2019communication}.  In \cite{chen2018communication}, Lazily
Aggregated Policy Gradient (LAPG), a distributed PG algorithm is developed,  which can reduce  the communication rounds between the agents and a central controller. This is achieved  by judiciously designing communication trigger rules. In \cite{ren2019communication},  the same policy evaluation problem as \cite{wai2018multi} is addressed, and develops a   hierarchical distributed algorithm by proposing a mixing matrix different from the doubly stochastic one used in \cite{zhang2018fully,wai2018multi,lee2018primal}, which  saves  communication  by allowing unidirectional information exchange among agents. In comparison, \cite{lin2019communication} proposes 
a distributed actor-critic algorithm, which reduces the communication by transmitting only   one scalar entry of its state vector at each iteration. The same   convergence guarantee as \cite{zhang2018fully} can be established.   

We note that RL under this decentralized setting with networked agents  has been studied beyond the multi-agent setting. Indeed, several works have modeled the setting for \emph{multi-task} RL, where multiple cooperative agents are also connected by a communication network, without any coordination from a central controller. However, each agents is in face of an \emph{independent MDP}, which is not influenced by other agents. Different agents may still have different reward functions, while the goal is to learn the optimal joint policy that optimizes the long-term return corresponding to the team-average reward.   In some sense, this setting can be deemed as a simplified version of the our MARL setting, for the less coupling among agents.  Under this setting, 
\cite{pennesi2010distributed} develops  a distributed actor-critic algorithm, where each agent first conducts a local TD-based critic step, followed by a consensus-based actor step that calculates the gradient based on the neighbors' information exchanged.  The  gradient of the average return is then shown to converge to zero.  In \cite{macua2017diff},  \emph{Diff-DAC}, another distributed actor-critic algorithm is developed  from duality theory. The updates, which  look similar to those of \cite{zhang2018fully},  are essentially  an example  of
the dual ascent method to solve  some  linear program. This provides   additional insights into the  actor-critic update for this setting.

Policy evaluation has also been considered under this setting of networked agents interacting with \emph{independent} MDPs.   The early work \cite{macua2015distributed} studies off-policy evaluation using  the importance sampling technique. Without coupling among agents, there is no need for 
each  agent   to know the actions of the others. 
Then, a 
 diffusion-based distributed gradient-TD method is proposed, which is proven to converge  with a sublinear rate in the mean-square sense. \cite{stankovic2016multi} then proposes two other variants of the gradient-TD updates, i.e., GTD2 and TDC \citep{sutton2009fast}, and proves   weak convergence using  the general stochastic approximation theory developed  in \cite{stankovic2016distributed}. 
 \cite{stankovic2016multi} specifically  considers the case where   agents are connected by a time-varying communication network. 
 The aforementioned work  \cite{cassano2018multi} also considers the independent MDP setting, with the same results established as the actual MARL one.  

%mention the other related research, e.g., comm. efficient., under attack/with robustness, 

%\begin{itemize}
%%	\item Independent Q learning works in this case, with convergence.
%%\item Learning in Markov potential games \cite{valcarcel2018learning}, survey or modeling of MPG \cite{gonzalez2013discrete,zazo2016dynamic} 
%\item MARL with Networked agents: Q-learning \cite{kar2013cal};  Actor-Critic \cite{zhang2018fully}, and all follow-ups, including \emph{asymptotic} \cite{lee2018primal}  and \emph{finite-time} analyses for policy evaluation \cite{wai2018multi,cassano2018multi,doan2019finitea,doan2019finiteb}, and actor-critic \cite{suttle2019multi} and consensus on actor   \cite{zhang2019distributed}. 
%%\item RNN / LSTM-based method (DRQN)
%%\item Learning to communicate and its followup work
%\item Cooperative network/graphical games: \cite{zhang2018model} 
%\item Communication-efficient MARL \cite{chen2018communication,ren2019communication,lin2019communication}
%%\item Mean-field control / swarm system (a large number of homogeneous agents), or mean-field teams.
%%\item Actor-critic fictitious play for \emph{multi-stage} MARL \cite{perolat2018actor}. 
%%\item Fictitious play also works for cooperation, namely, is shown to converge for cooperative, potential (normal-form) games \cite{hofbauer2002global}.
%\end{itemize}

\section{Concluding Remarks}\label{sec:conclusion}

Owing to the ubiquity of sequential decision-making in presence of more than one agents, multi-agent RL has long been a significant  while challenging research area. In this review, we have summarized the recent advances in a sub-area of MARL: decentralized MARL with networked agents. Particularly, we have focused on the MARL algorithms that concern this setting, and are backed by theoretical analysis. We hope our review is appealing to the researchers of similar interests, and has provided stimulus for them to continue pursuing this direction. Interesting while open future  directions may concern the setting with partial observability, with adversarial agents in the system. It is also interesting  to  develop theoretical results for MARL algorithms under this setting with deep neural networks as function approximators, which have already achieved tremendous empirical success. 
See \cite{zhang2019multiagent} for more discussions on intriguing future directions. 
\bibliographystyle{fitee}
\bibliography{MARL_Springer_1,MARL_Springer_2}

\end{document}